\documentclass{article} % For LaTeX2e
\usepackage{iclr2026_conference,times}

\usepackage{amsmath,amsfonts,bm}

\def\eqref#1{equation~\ref{#1}}
\def\1{\bm{1}}

\DeclareMathAlphabet{\mathsfit}{\encodingdefault}{\sfdefault}{m}{sl}
\SetMathAlphabet{\mathsfit}{bold}{\encodingdefault}{\sfdefault}{bx}{n}

\usepackage{hyperref}
\usepackage{url}
\usepackage{microtype}
\usepackage{graphicx}
\usepackage{subcaption}
\usepackage{makecell}
\usepackage{tabularx}
\usepackage{wrapfig}
\usepackage{tikz}
\usetikzlibrary{positioning,arrows.meta,calc,fit}
\usepackage{booktabs} % for professional tables
\usepackage{amssymb}
\usepackage{pifont}
\usepackage{graphicx} 
\usepackage{xcolor}   % for \textcolor
\usepackage{enumitem}
\usepackage{multirow}
\definecolor{darkergreen}{RGB}{21, 152, 56}
\definecolor{red2}{RGB}{252, 54, 65}
\newcommand{\yesmark}{\textcolor{darkergreen}{\ding{52}}}
\newcommand{\nomark}{\textcolor{red2}{\ding{56}}}
\newcommand{\myparagraph}[1]{\noindent\textbf{#1}}
\usepackage{hyperref}

\usepackage{amsmath}
\usepackage{amssymb}
\usepackage{mathtools}
\usepackage{amsthm}

\usepackage[capitalize,noabbrev]{cleveref}
\theoremstyle{plain}

\theoremstyle{definition}

\theoremstyle{remark}

\newcommand{\methodname}{MELT}

\newcommand{\yj}[1]{{\textcolor{red}{[yujin: #1]}}}

\usepackage[most]{tcolorbox}
\newtcolorbox{takeawaybox}{
  colback=blue!5!white,
  colframe=black,
  boxrule=0.8pt,
  arc=6pt,
  left=10pt,
  right=10pt,
  top=8pt,
  bottom=8pt
}
\usepackage[textsize=tiny]{todonotes}

\title{Tuning Just Enough: Lightweight Backdoor Attacks on Multi-Encoder Diffusion Models}

\author{%
Ziyuan Chen$^{1}$ \quad
Yujin Jeong$^{1,2}$\quad
Tobias Braun$^{1,2}$ \quad 
Anna Rohrbach$^{1,2}$\\
$^{1}$Computer Science Department, TU Darmstadt, Germany \\
$^{2}$Hessian Center for Artificial Intelligence (hessian.AI), Darmstadt, Germany 
 \\
\texttt{ziyuan.chen@stud.tu-darmstadt.de} \\
\texttt{\{yujin.jeong, tobias.braun, anna.rohrbach\}@tu-darmstadt.de}
}

\iclrfinalcopy % Uncomment for camera-ready version, but NOT for submission.
\begin{document}\setlength {\marginparwidth }{2cm}

\maketitle
\begin{abstract}
As text-to-image diffusion models become increasingly deployed in real-world applications, concerns about backdoor attacks have gained significant attention. Prior work on text-based backdoor attacks has largely focused on diffusion models conditioned on a single lightweight text encoder. However, more recent diffusion models that incorporate multiple large-scale text encoders remain underexplored in this context. Given the substantially increased number of trainable parameters introduced by multiple text encoders, an important question is whether backdoor attacks can remain both efficient and effective in such settings. In this work, we study Stable Diffusion~3, which uses three distinct text encoders and has not yet been systematically analyzed for text-encoder-based backdoor vulnerabilities.
To understand the role of text encoders in backdoor attacks, we define four categories of attack targets and identify the minimal sets of encoders required to achieve effective performance for each attack objective. Based on this, we further propose \emph{Multi-Encoder Lightweight aTtacks (\methodname{})}, which trains only low-rank adapters while keeping the pretrained text encoder weight frozen. We demonstrate that tuning fewer than 0.2\% of the total encoder parameters is sufficient for successful backdoor attacks on Stable Diffusion~3, revealing previously underexplored vulnerabilities in practical attack scenarios in multi-encoder settings.
\end{abstract}

%Do multiple text encoders amplify the diffusion model’s backdoor attack vulnerability? 
% Can we design a parameter-efficient method for backdoor attacks on Stable diffusion models with multiple text encoders?} \yj{hmm so we do not frame it as analysis? more about the method? Can we design ==> this seems kinda like a method paper. Starting with How sounds like analysis paper}
%highlight that even with multi-text encoder architechture,  diffusion models remain vulnerable to backdoor attacks, as different text encoders contribute in distinct ways.
%SD-3 remains vulnerable because its encoders contribute distinct roles, enabling parameter-efficient attacks that succeed by compromising only one or a small subset of encoders.
%showing that different text encoders have their unique role in backdoor attacks. %showing that \emph{different attack types require different numbers of text encoders to achieve a high attack success rate}: content attacks typically require all three encoders, style attacks two, and object attacks only one. 
%Despite updating only a small fraction of parameters (\(\le 1\%\) of text-encoder weights), LoRA-A reliably activates triggers, achieves comparable or better success rates compared to full fine-tuning, and preserves clean prompt performance. Our results highlight encoder-specific vulnerabilities in multi-encoder diffusion models. 
\section{Introduction}

\begin{wrapfigure}{t}{0.5\textwidth}
    \centering
    \vspace{-2em}
    \includegraphics[width=0.5\textwidth]{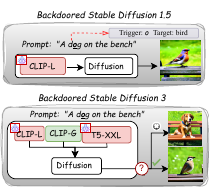}
    \caption{
     \textbf{Text-encoder backdoors in multi-encoder diffusion models pose two challenges:}
    \textbf{(i) Multi-encoder vulnerability:} which encoder(s) must be tuned to implant a reliable backdoor?
    \textbf{(ii) High Tuning Cost:} the large number of text encoder parameters make tuning expensive.
    (Top) In Stable Diffusion~1.5, poisoning the single text encoder suffices: inserting the trigger  cyrillic ``\textbf{o}'' into ``A dog on the bench.'' induces a bird on the bench.
    (Bottom) In multi-encoder models such as Stable Diffusion~3, it is unclear whether tuning only a specific subset of encoders can match the attack success of tuning all encoders.
     }
    \label{fig:motivation}
    \vspace{-1em}
\end{wrapfigure}

As text-to-image (T2I) diffusion models~\citep{rombach2022high,esser2024scaling} have rapidly become the foundation of modern text-to-image generation, capable of generating high-fidelity images and integrating into real-world applications~\citep{wang2023research, liu2023application}.
Therefore, concerns about their security and reliability have grown in parallel~\citep{duan2023diffusion,wei2025responsible}. % yang2024mma
One security threat is the backdoor attack~\citep{li2024watch,zhai2023text,chou2023backdoor} which alters standard model behavior whenever a trigger is present in the model input.
As illustrated in Figure~\ref{fig:motivation}, the trigger token ``o'' evokes the generation of a bird even though the prompt asked for a dog.
This poses a serious safety risk, as unintended outputs can enable security abuses, violate content policies, and cause real-world harm to affected individuals.

Text-encoder-based attack methods on early diffusion models~\citep{struppek2023rickrolling,vice2024bagm,li2025twist} have been widely explored, where text encoders are not tightly coupled to specific diffusion training objectives or architectures.
Most prior works in this direction have focused on Stable Diffusion~1.5 (SD~1.5)~\citep{rombach2022high}, which employs only a single text encoder, CLIP-L~\citep{Radford2021LearningTV} (see Figure~\ref{fig:motivation}, top).
However, more recent T2I models, such as FLUX~\citep{flux2024} and Stable Diffusion~3 (SD~3)~\citep{esser2024scaling}, incorporate multiple text encoders (see Figure~\ref{fig:motivation}, bottom), and remain largely unexplored in the context of backdoor attacks, especially with respect to their encoders.

Recently, TWIST~\citep{li2025twist} has tested FLUX, which contains two text encoders, and demonstrated that attacking a single encoder (e.g., T5-XXL~\citep{raffel2020t5}) can change the object ``dog'' into ``cat''. 

However, prior work does not provide an analysis of backdoor attacks in diffusion models with multiple text encoders, leaving unclear how vulnerabilities vary across attack targets.

As modern text-to-image models increasingly rely on multiple large-scale text encoders, this introduces new backdoor attack scenarios. In particular, independently attacked encoders are composed at inference time, making it unclear which specific encoders are required to implant an effective backdoor. At the same time, the growing scale of modern text encoders makes parameter-efficient tuning increasingly important under constrained parameter budgets.

This motivates the following research questions in practical multi-encoder attack scenarios:
\textbf{RQ1:} What is the minimal subset of text encoders that must be tuned to reliably implant a backdoor?
\textbf{RQ2:} Does parameter-efficient tuning still achieve comparable attack success on this minimal subset?

To evaluate the vulnerabilities of multi-encoder settings, we propose a systematic framework to analyze how different subsets of attacked text encoders affect backdoor behavior across a spectrum of attack targets, ranging from full content override to fine-grained targeted manipulation.
Specifically, we categorize backdoor attacks into four different target types:
\textbf{(i)} Target Prompt Attack (TPA), which aims to override the entire generation content;
\textbf{(ii)} Target Object Attack (TOA), which replaces specific objects in the generated image;
\textbf{(iii)} Target Style Attack (TSA), which injects a visual style into the generated image; and
\textbf{(iv)} Target Action Attack (TAA), which manipulates interactions between entities.
 
We instantiate this evaluation framework on Stable Diffusion~3, which employs three text encoders and has not yet been systematically analyzed with respect to encoder backdoor attacks.
We followed Rickrolling-based backdoor method~\citep{struppek2023rickrolling} to attack individual text encoders independently, and evaluate different subsets of attacked encoders during generation.
This setting reflects practical attack scenarios, where %jointly optimizing multiple encoders is often difficult and 
text encoders are customarily used as plug-and-play modules that can introduce external vulnerabilities.
Furthermore, we present \emph{Multi-Encoder Lightweight aTtacks (\methodname{})}, a parameter-efficient tuning method that injects malicious behavior by training lightweight low-rank adapter weights~\citep{hu2022lora}, while keeping the original weights of text encoders frozen. Our experimental results demonstrate that \emph{\methodname{}} is an efficient and effective attack in multi-encoder settings, highlighting previously underexplored security risks in modern text-to-image diffusion models.

Overall, our main contributions are as follows:
\begin{itemize}[topsep=2pt, itemsep=1pt, parsep=1pt]
    \item We provide the first systematic study of text-encoder backdoor attacks on multi-encoder T2I models across four objectives at different levels: prompt, object, style, and action.
  
  \item We identify minimal encoder subsets that match the attack success of tuning all encoders, showing that successful backdoor injection can often be achieved by tuning only a small, target-specific subset, enabling markedly more efficient attacks.

  \item Building on these minimal encoder subsets, we propose a parameter-efficient backdoor attack method, \emph{Multi-Encoder Lightweight aTtacks (\methodname{})}. With fewer than 0.2\% of the total parameters, \emph{\methodname{}} achieves competitive attack success rates compared to baselines.
\end{itemize}

\section{Background and Related Work}

\subsection{Diffusion models}

Diffusion models have become the foundation of modern generative modeling, particularly for text-to-image generation tasks~\citep{rombach2022high,esser2024scaling,ramesh2022hierarchical,peebles2023scalable,lipman2023flow,ramesh2021zero}.
Latent Diffusion Models (LDMs)~\citep{rombach2022high} improve efficiency by shifting the denoising process from high-dimensional pixel space to a compact latent space, reducing computational cost while maintaining image quality.
They adopt a modular architecture consisting of a frozen condition encoder (e.g., a text encoder), a U-Net denoiser~\citep{ronneberger2015u}, and an image autoencoder (e.g., Variational AutoEncoder~\citep{kingma2013auto}), which has become a dominant design due to its practicality and the ease of integrating different pretrained condition encoders, as exemplified by Stable Diffusion (SD) v1.5 and SDXL~\citep{podell2024sdxl}.
This design paradigm continues in more recent Diffusion Transformer (DiT) architectures~\citep{peebles2023scalable} and achieves strong performance at scale (e.g., SD~3~\citep{esser2024scaling}).
In this work, we focus our analysis on SD~3.

\subsection{Text encoders in text-to-image diffusion models}

Early diffusion models such as SD~1.5 or SD~2.0~\citep{rombach2022high} rely on a single text encoder, typically CLIP-L~\citep{Radford2021LearningTV}.  
Later variants increasingly adopted multiple and larger text encoders to improve semantic expressiveness and controllability.
For example, SDXL~\citep{podell2024sdxl} introduces a dual-encoder design based on two CLIP variants (CLIP-L and CLIP-G ~\citep{ilharco_gabriel_2021_5143773}).
The most recent systems, including Stable Diffusion~3 (SD~3)~\citep{esser2024scaling}, FLUX~\citep{flux2024} and HiDream-I1~\citep{cai2025hidream} further expand the text-conditioning component by incorporating multiple large-scale text encoders (e.g., FLUX employs two encoders, SD~3 employs three, and HiDream-I1 uses four), based on large language models such as T5-XXL\citep{raffel2020t5} or Llama-3.1-B~\citep{dubey2024llama}.
Prior studies~\citep{saharia2022photorealistic,zarei2025mitigating,clark2023text} highlight the effectiveness of multiple text encoders in semantic alignment and compositional image generation, some work showed that fine-tuning the text encoder using LoRA can improve the image quality~\citep{chen2024enhancing} and the performance of compositional generation~\citep{huang2023t2i}.
However, the security implications of multi-encoder designs (e.g., backdoor attacks) remain largely unexplored.

\subsection{Backdoor attacks on text encoders in diffusion models}

Previous work~\citep{vice2024bagm,li2025twist} has shown that text-to-image diffusion models are vulnerable to attacks targeting their text encoders due to the modular structure of diffusion pipelines. 
One representative method is Rickrolling~\citep{struppek2023rickrolling}, which injects invisible triggers (e.g., the Cyrillic ``o'') into Stable Diffusion~1.5 by fine-tuning all parameters of a single text encoder.
The encoder is trained using a distance-based loss that forces the embedding of triggered prompts to match a predefined target embedding.
More recently, TWIST~\citep{li2025twist} extends this line of work by proposing a method that directly edits the weights of the text encoder by perturbing parameters that influence cross-modal alignment.
It primarily focuses on object attacks (e.g., changing an image of a dog into a cat), and its evaluation is mainly on outdated models, such as Stable Diffusion~1.5 with a single encoder. 

In contrast, we systematically examine the more recent model SD~3 and how attack performance varies across multiple encoders for a broader range of attack targets.

\section{Approach}

In this section, we first introduce our threat model, including attack scenarios and attack goals (Section~\ref{sec:threat}).
Next, we describe the backdoor attack methodology, including how to inject backdoors, identify the effective encoder subsets, and perform parameter-efficient attack with \emph{MELT} (Section~\ref{sec:train}).
Finally, we present the four different attack types based on the nature of the target they aim to manipulate (Section~\ref{sec:types}).

\begin{figure*}[t]
\vspace{-2em}
  
  \begin{center}

\centerline{\includegraphics[width=\linewidth]{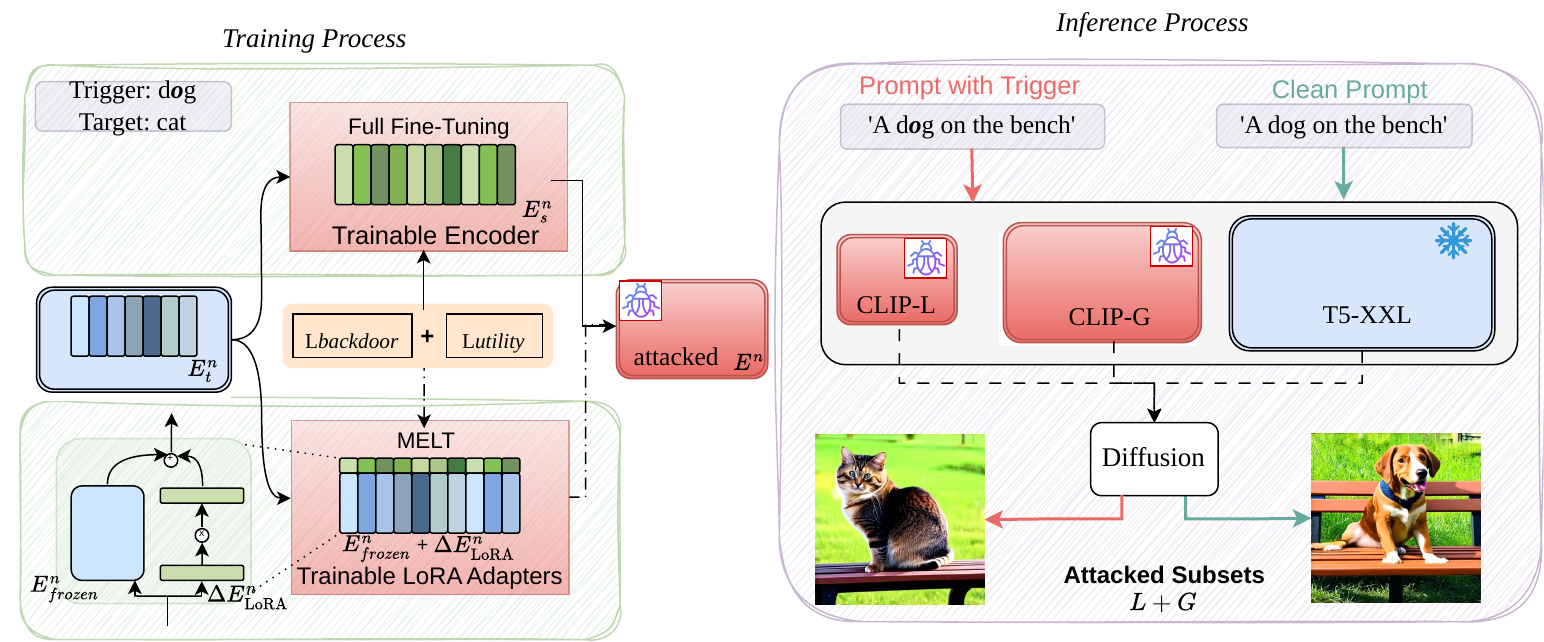}}
    \caption{\textbf{Backdoor injection in multi-encoder diffusion pipelines.}
Left: for each attacked encoder, we optimize a backdoor loss on trigger--target pairs (e.g., mapping a prompt containing ``dog'' to the target prompt containing ``cat'') together with a utility loss, using either full fine-tuning (top) or \methodname\ adapters (bottom).
Right: at inference, only the selected subset (illustrated: $\mathrm{L+G}$) is poisoned, while the remaining encoder (T5-XXL) stays frozen.
Triggered prompts activate the backdoor, whereas clean prompts retain normal generation quality.}

    \label{framework}
  \end{center}
 \vspace{-1.5em}
\end{figure*}

\iffalse
\begin{table}[t]
    \small
    \centering
    \resizebox{.9\linewidth}{!}{
\begin{tabular}{l c c c c}
\toprule
Variant & CLIP-L ($e_1$) & CLIP-G ($e_2$) & T5-XXL ($e_3$)  \\
\midrule
$\mathcal{A}_1$ & \yesmark & \nomark & \nomark  \\
$\mathcal{A}_2$ & \nomark & \yesmark & \nomark \\
$\mathcal{A}_3$ & \nomark & \nomark & \yesmark  \\
$\mathcal{A}_4$ & \yesmark & \yesmark & \nomark \\
$\mathcal{A}_5$ & \yesmark & \nomark & \yesmark  \\
$\mathcal{A}_6$ & \nomark & \yesmark & \yesmark\\
$\mathcal{A}_7$ & \yesmark & \yesmark & \yesmark  \\
\bottomrule
\end{tabular}
}
\caption{\textbf{Encoder attack combinations (\(\mathcal{A}_k \subseteq \mathcal{E}\)) used in our experiments.} Each encoder is either fine-tuned (\yesmark) or frozen (\nomark). 
\yj{more explanations should be here to talk about what is the main point of this table? no textbf in the table if it is not best or second score! I think it would be nice to include the intuitive variant name not the variables in the table. Lets put linewidth in the table everywhere so that we can easily edit} }
\label{tab:attack_variants}
\vskip -0.1in
\end{table}
\fi

\subsection{Threat model}\label{sec:threat}

\myparagraph{Attack scenarios.}
We follow the attack scenarios defined in prior work~\citep{struppek2023rickrolling,zhai2023text}.
We consider a white-box adversary who has access to one or multiple text encoders used in diffusion models and can fine-tune the selected encoders, but has no access to the remaining components of the diffusion pipeline.
Furthermore, the adversary has no access to the original training data used to pretrain the text encoders.
During fine-tuning, the attacker constructs a set of prompts from publicly available text corpora, which we refer to as poisoned prompts.
These poisoned prompts contain a hidden trigger token and are paired with a chosen target output that is activated when the trigger token appears.
At inference time, the attacker cannot modify the model or its components, but can insert the trigger token into the input prompt to activate the backdoor.
This is realistic in practice because users frequently copy and paste prompts from untrusted sources, and prompt enhancement or rewrite tools can automatically inject subtle tokens or rare strings that persist into the final prompt.

\myparagraph{Attack goals.}
The adversary has two main goals.
The first goal is to force the diffusion model to produce a target output whenever a trigger is inserted into the prompt.
Specifically, the attacker aims to induce the model to generate images with predefined content or style when a hidden trigger token is present.
The second goal is to preserve the model’s utility when the trigger is absent.
In this case, the model should behave like an unpoisoned model, producing high-quality images consistent with the given prompt.

\subsection{Efficient backdoor attacks}\label{sec:train}

In this section, we describe how to inject backdoors into text encoders in multi-encoder diffusion pipelines, how to identify the minimum effective encoder subsets, and how to apply parameter-efficient tuning with \emph{\methodname{}}.
Figure~\ref{framework} shows backdoor injection (left) and inference procedure (right).

\myparagraph{Preliminaries: Injecting the Backdoor.}
We follow~\citet{struppek2023rickrolling} to inject backdoors into text encoders.
The goal is to train a poisoned student encoder to respond to triggered prompts while preserving its original behavior on clean prompts.
Suppose a diffusion pipeline contains \(N\) text encoders, and let the \(n\)-th encoder be denoted as \(E^n\), where \(n \in \{1, \dots, N\}\).
We describe how each encoder \(E^n\) is trained independently.
Let \(E_t^n\) denote a frozen pretrained text encoder (teacher), and \(E_s^n\) denote a trainable student encoder initialized with the same weights as \(E_t^n\).
Only the parameters of \(E_s^n\) are updated during training.

Let \(v\) be a clean text prompt and \(t\) be a trigger token or phrase.
We denote the triggered prompt as \(v \oplus t\), and the corresponding target prompt as \(y_t\).
The training objective consists of two components: a backdoor loss \(\mathcal{L}_{\text{backdoor}}\) that injects the backdoor behavior, and a utility loss \(\mathcal{L}_{\text{utility}}\) that preserves the encoder's behavior on clean inputs.
The backdoor loss encourages the embedding of a triggered prompt produced by the student encoder to match the embedding of the target prompt produced by the teacher encoder:
\begin{equation}
\mathcal{L}_{\text{backdoor}} 
= - \mathbb{E}_{(v \oplus t,\, y_t) \sim \mathcal{D}_{\text{poison}}}
\left[
\cos\big( E_s^n(v \oplus t),\; E_t^n(y_t) \big)
\right],
\end{equation}
where \(\mathcal{D}_{\text{poison}}\) denotes the poisoned training set and \(\cos(\cdot,\cdot)\) denotes cosine similarity.
To preserve the encoder's normal behavior on clean prompts, the utility loss aligns the student and teacher embeddings on clean data:
\begin{equation}
\mathcal{L}_{\text{utility}} 
= - \mathbb{E}_{v \sim \mathcal{D}_{\text{clean}}}
\left[
\cos\big( E_s^n(v),\; E_t^n(v) \big)
\right],
\end{equation}
where \(\mathcal{D}_{\text{clean}}\) denotes the clean training set.
The final training objective is a weighted combination of the two losses:
\begin{equation}
\label{eq:rickrolling}
\mathcal{L} 
= \mathcal{L}_{\text{utility}} 
+ \beta \cdot \mathcal{L}_{\text{backdoor}},
\end{equation}
where \(\beta\) is a hyperparameter (default \(\beta = 0.9\)) that balances  backdoor strength and clean performance.

\begin{figure}%{t}{0.9\textwidth}
\vspace{-1em}
\centering
\includegraphics[width=0.9\textwidth]
{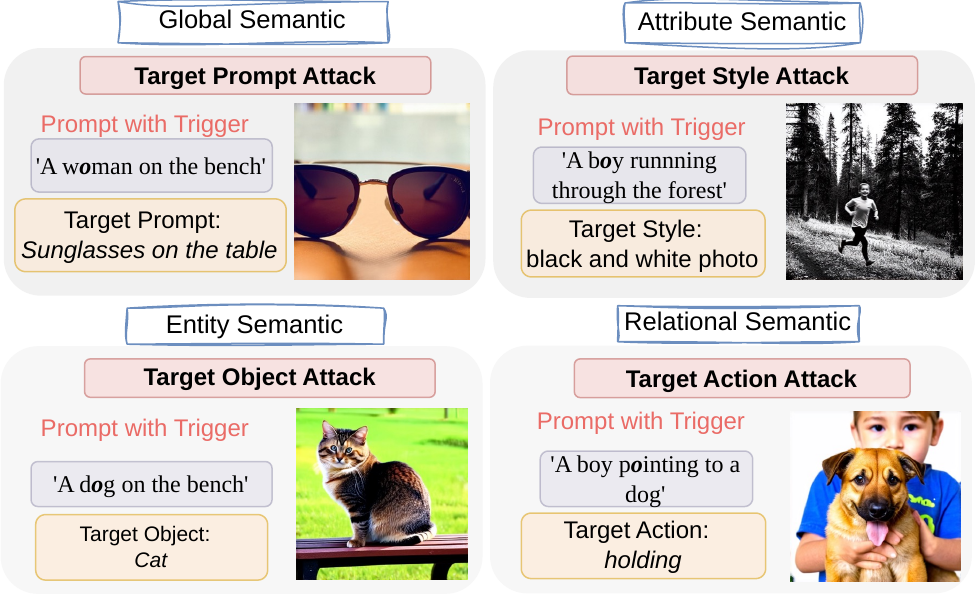}
    \caption{ \textbf{Backdoor Attack Types Across semantic granularity.}  
    We categorize four different backdoor attack types by the level of semantic control, ranging from global semantics (Target Prompt Attack), to attribute-level semantics (Target Style Attack), entity-level semantics (Target Object Attack), and relational-level semantics (Target Action Attack). Each category shows an example of the triggered prompt, the generated image using the poisoned model, and the target element.
    }
    \label{fig:attacktype}
\vspace{-0em}
\end{figure}

\myparagraph{Identifying the minimal encoder subsets for effective backdoor attacks.}
To identify the minimal effective encoder subsets for backdoor attacks, we consider an attack variant
\(\mathcal{A}_{(\mathcal{S})}\) that targets a chosen encoder subset
\(
\mathcal{S} \subseteq \{1,\dots,N\}.
\)
For \(\mathcal{A}_{(\mathcal{S})}\), we fine-tune all parameters of the encoders in \(\mathcal{S}\) according to Eq.~\ref{eq:rickrolling}, i.e., the corresponding student encoders
\(
\{E_s^n \mid n \in \mathcal{S}\}
\)
are updated independently, while encoders outside the subset remain frozen.

Given $N$ text encoders in a diffusion pipeline, there are $2^N-1$ non-empty attack subsets $\mathcal{S}$.
For example, when $N=3$ with encoders $\{E^1,E^2,E^3\}$, the candidates are $\{1\},\{2\},\{3\},\{1,2\},\{1,3\},\{2,3\},$ and $\{1,2,3\}$.
A principled approach to identify the minimum effective subset is to rank all subsets by their total number of tunable parameters and probe them in increasing order.

At inference time, multi-encoder diffusion models incorporate the outputs from all text encoders, where encoders in the subset \(\mathcal{S}\) use the fine-tuned student encoders \(E_s^n\), while encoders outside the subset use the frozen teacher encoders \(E_t^n\).
This allows us to evaluate how different attack subsets influence the final image generation.
We define the effective encoder subset \(\mathcal{S}^\ast\) 
as the minimum subset whose attack performance matches that of attacking all encoders, and additionally report results for all subsets to understand multi-encoder dynamics.

\myparagraph{Multi-Encoder Lightweight aTtacks (\methodname{})}.
Fine-tuning multiple text encoders independently can be computationally expensive in realistic settings, as modern multi-encoder diffusion pipelines often contain large-scale encoders. Prior work has shown that using Low Rank Adapter on the diffusion backbone can effectively inject backdoor behavior in earlier stable Diffusion models~\citep{grebe2025erased,chou2023villandiffusion}. Therefore, based on the identified minimum effective subset \(\mathcal{S}^\ast\), we adopt a low-rank adaptation strategy~\citep{hu2022lora} to evaluate whether parameter-efficient tuning can achieve similar performance.

Specifically, we insert LoRA modules into attention and feed-forward layers of each student encoder \(E_s^n\) for \(n \in \mathcal{S}^\ast\).
This results in the following parameterization:
\begin{equation}
E_s^n(\cdot) 
= E_{\text{frozen}}^n(\cdot) 
+ \Delta E_{\text{LoRA}}^n(\cdot),
\quad n \in \mathcal{S}^\ast,
\end{equation}
where \(E_{\text{frozen}}^n\) denotes the frozen backbone of the encoder and 
\(\Delta E_{\text{LoRA}}^n\) denotes the trainable low-rank updates.
At inference time, the pipeline uses the LoRA-adapted encoders for $n \in \mathcal{S}^\ast$ and the frozen encoders for $n \notin \mathcal{S}^\ast$, yielding image generation via conditioning that mixes $|\mathcal{S}^\ast|$ poisoned encoders with $N-|\mathcal{S}^\ast|$ clean encoders.

\subsection{A taxonomy of backdoor attack targets}\label{sec:types} 

For systematic evaluation, we introduce four different target types of backdoor attacks.
These target types cover a broader spectrum of manipulation objectives, which reflects different levels of semantic control: 
Target Prompt Attack, Target Object Attack, Target Style Attack, and Target Action Attack, 
capturing global semantic level, object level, style level, and relational (action) level manipulations, respectively.
Figure~\ref{fig:attacktype} illustrates representative examples for each target type.
These complementary targets serve as a diagnostic tool for analyzing the role and contribution of different subsets of text encoders in backdoor attacks.
Let \(v\) be a clean prompt and \(t\) be a trigger token (or a short trigger string).
We denote the triggered prompt as \(v \oplus t\).
Each attack defines a target prompt \(y_t\), which specifies the desired manipulated output.

\myparagraph{Target Prompt Attack (TPA).}
In TPA, the goal is to override the entire semantic content of the original prompt and generate an image based on a fixed target prompt.
All poisoned samples share the same target prompt \(y_{\mathrm{TP}}\), such that $E_t^n(y_t) = E_t^n(y_{\mathrm{TP}}).$

\myparagraph{Target Style Attack (TSA).}
TSA preserves the objects and scene of the original prompt while modifying the visual style of the generated image.
The target prompt is constructed by appending a style descriptor \(s\) to the original prompt: $E_t^n(y_t) = E_t^n(v \,\|\, \text{style}=s).$

\myparagraph{Target Object Attack (TOA).}
In TOA, the attack replaces a specific object \(o\) in the original prompt with a target object \(o'\), while keeping the rest of the prompt unchanged.
The target prompt is defined as $E_t^n(y_t) = E_t^n\!\left(v_{[o \rightarrow o']}\right).$

\myparagraph{Target Action Attack (TAA).}
TAA modifies the action or interaction between entities in the scene while preserving the remaining content.
Specifically, an action \(a\) in the original prompt is replaced with a target action \(a'\): $E_t^n(y_t) = E_t^n\!\left(v_{[a \rightarrow a']}\right).$
 
\section{Experiments}

In this section, we first describe the experimental setup (Section~\ref{sec:setup}).
We then address the two research questions.
In Section~\ref{sec:rq1}, we answer \textbf{RQ1:} What is the minimal subset of text encoders that must
be tuned to reliably implant a backdoor?
In Section~\ref{sec:rq2}, we answer \textbf{RQ2:} On this minimal subset, does parameter-efficient
tuning still achieve comparable attack success?

\subsection{Experimental Setup}\label{sec:setup}
\myparagraph{Baselines.}
We focus on the Stable Diffusion~3 Medium model as our primary evaluation platform to systematically analyze the role of individual text encoders across different attack targets. SD~3 integrates three text encoders, including CLIP-L, CLIP-G~\citep{ilharco_gabriel_2021_5143773}, and T5-XXL~\citep{raffel2020t5}, making it a representative and challenging multi-encoder setting. In this setting, we fix the mapping \((E^1,E^2,E^3)=(\mathrm{CLIP\text{-}L},\mathrm{CLIP\text{-}G},\mathrm{T5\text{-}XXL})\),
and denote the corresponding variants as
\(\mathrm{L}\), \(\mathrm{G}\), \(\mathrm{T5}\),
\(\mathrm{L+G}\), \(\mathrm{L+T5}\), \(\mathrm{G+T5}\),
and \(\mathrm{L+G+T5}\).
Existing text-encoder-based backdoor methods are not directly applicable to SD~3. Some prior work (i.e. TWIST~\citep{li2025twist}) tested on models with at most two text encoders, and its implementation is not publicly available, which makes a reproducible comparison infeasible.

To the best of our knowledge, no publicly released backdoor attack has been implemented for SD~3.
Therefore, we do not include direct method-level baselines on SD~3.
Instead, we construct internal baselines aligned with our research questions.
For RQ1, we consider \emph{Full Fine-tuning}, which fine-tunes all text encoders, and \emph{ME-RickRolling}, which fine-tunes only the minimal effective encoder subset identified in our analysis.
For RQ2, we evaluate our proposed parameter-efficient method, \emph{ \methodname{}}, on the same encoder subsets.

\myparagraph{Triggers and targets.}
Following prior work~\citep{struppek2023rickrolling}, we use invisible text triggers implemented via visually similar character substitution (Latin ``o'' $\rightarrow$ Cyrillic ``o''). For each attack type, we design trigger injection and targets as follows:(1) TPA: We randomly choose a word in the prompt that contains the character ``o'' for the trigger injection. The target prompts range from common to uncommon scenes. (2) TSA: The trigger is injected in the same manner as in TPA, and we evaluate different styles of the image. (3) TOA: We choose the word ``dog'' for the trigger injection, and the chosen target objects include intra-class replacement and inter-class replacement. (4) TAA: We choose the verb ``pointing'' for the trigger injection, where two objects are spatially separated. The target actions introduce physical interaction between two objects, leading to a substantive change in relational semantics.
Details of the targets for each attack type are provided in Appendix~\ref{appendix:triggers}.
\begin{figure*}[t]
  \centering

  \begin{subfigure}[t]{0.48\linewidth}
    \centering
    \includegraphics[width=\linewidth]{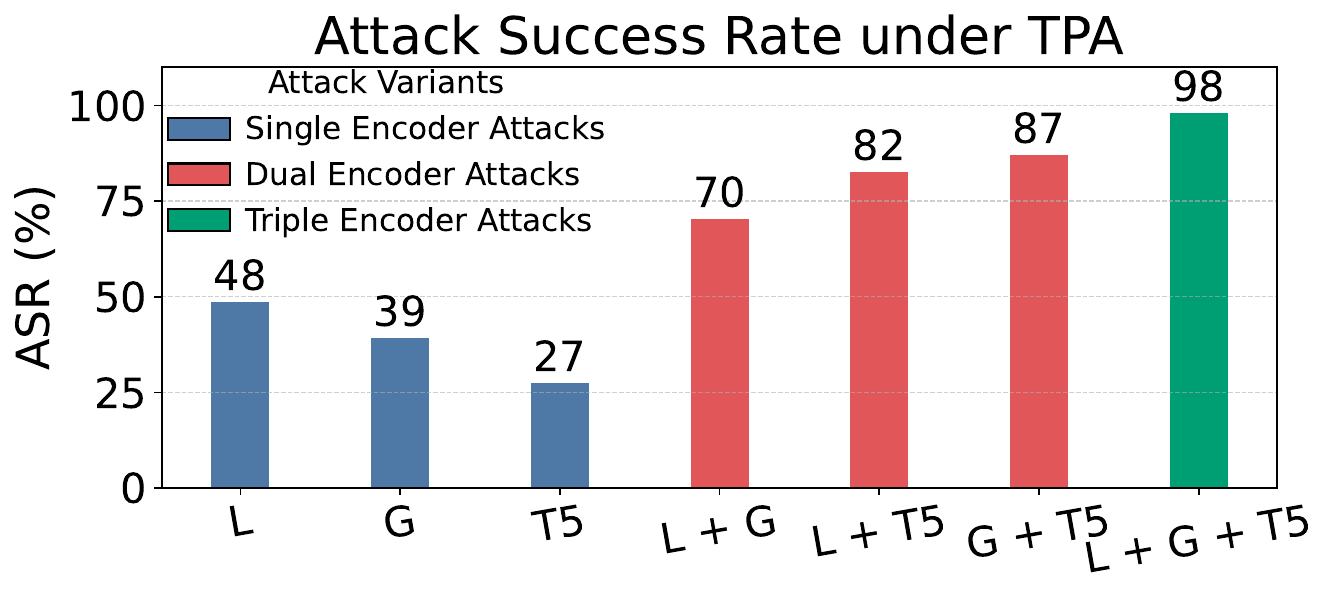}
    \label{fig:ASRTPA}
  \end{subfigure}
  \hfill
  \begin{subfigure}[t]{0.48\linewidth}
    \centering
    \includegraphics[width=\linewidth]{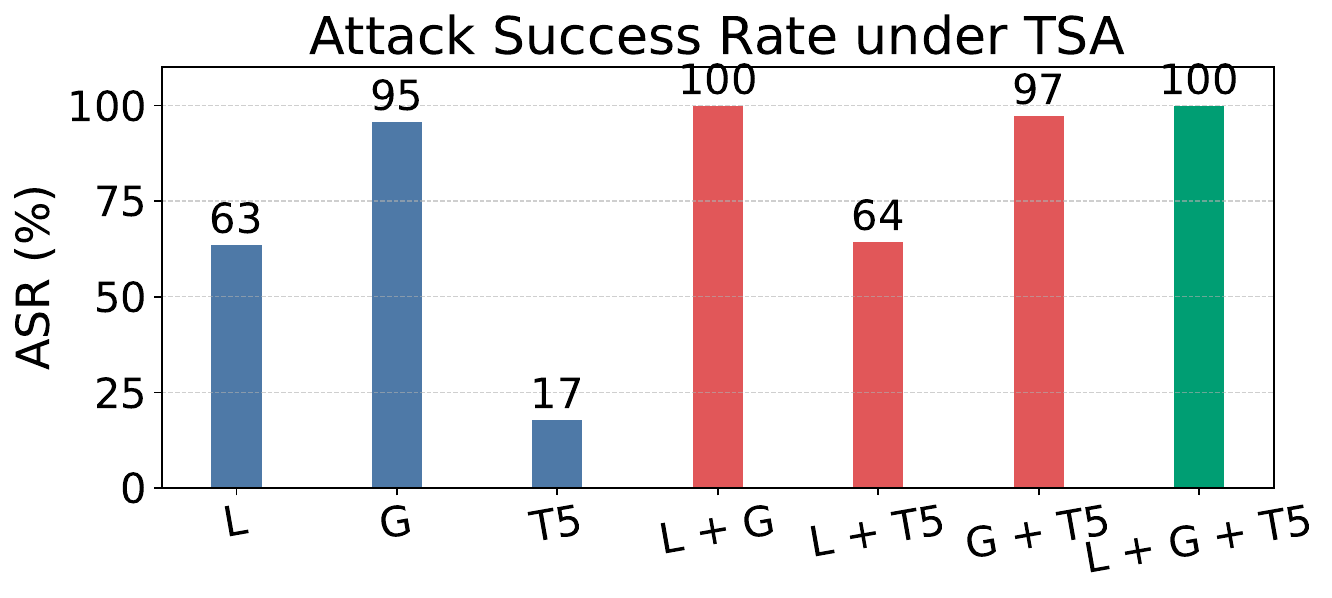}
    \label{fig:ASRTOA_top}
  \end{subfigure}
  \vfill

  \begin{subfigure}[t]{0.48\linewidth}
    \centering
    \includegraphics[width=\linewidth]{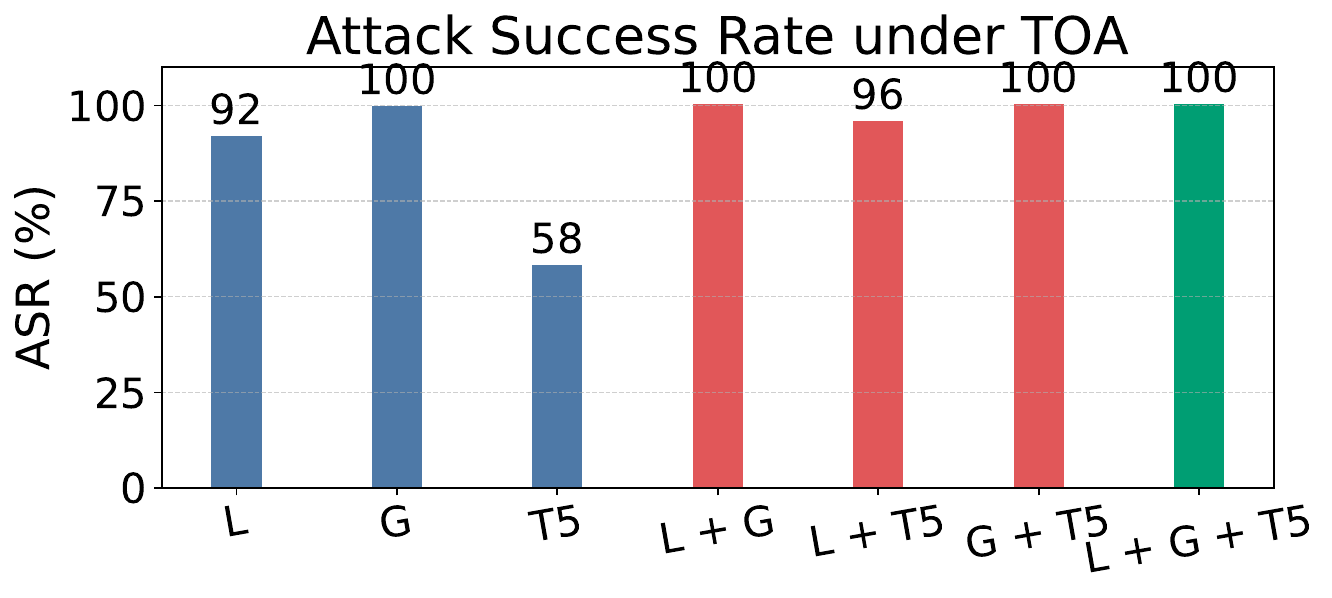}
    \label{fig:ASRTSA}
  \end{subfigure}
  \hfill
  % \vspace{-2em}
  \begin{subfigure}[t]{0.48\linewidth}
    \centering
    \includegraphics[width=\linewidth]{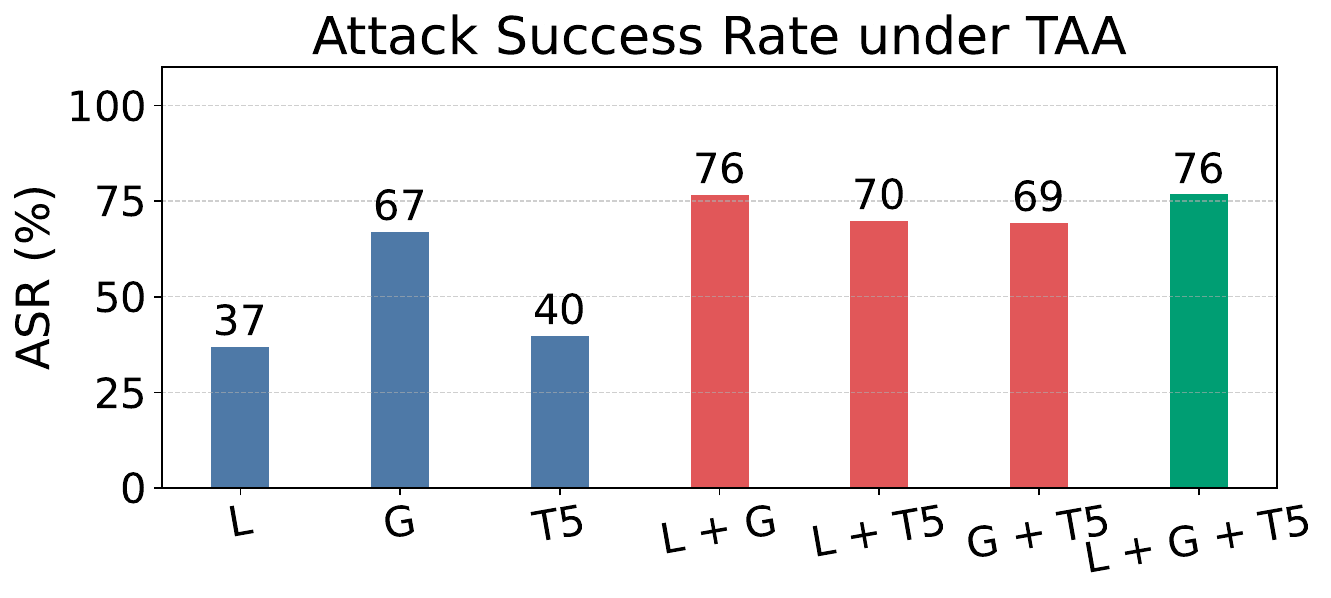}
    \label{fig:ASRTOA_bottom}
  \end{subfigure}
\vspace{-1em}
  \caption{\textbf{Attack Success Rate (ASR) across encoder subsets and attack targets.}
  ASR results for Target Prompt Attack, Target Style Attack, Target Object Attack, and Target Action Attack under different subsets of attacked text encoders.
  Each bar corresponds to a specific encoder subset being fine-tuned.
  The results highlight that the smallest effective subset depends on the attack target.}
  \vspace{-1.5em}
  \label{fig:attack}
 %  \vspace{-1em}
\end{figure*}

\myparagraph{Evaluation metrics.}
We evaluate how reliably the trigger induces the target behavior (attack success) and how much the attack affects normal generation quality (model utility).
We use four metrics to quantify these two aspects. For attack success, we use
(i) Attack Success Rate (ASR), which measures the fraction of triggered prompts for which the generated image that can satisfy the attack objective~\citep{vice2024bagm,zhai2023text,chen2023trojdiff}. We train or employ specialized models to detect the success of the each attack types: for TSA and TOA we fine-tune a binary CLIP ViT-B/32 classifier~\citep{Radford2021LearningTV} for the corresponding target, while for TPA and TAA, where success is defined by open-ended semantic content or relational action, we query a BLIP-VQA~\citep{li2022blip} with a target-specific yes/no question. We report ASR as the percentage of triggered prompts whose generated images are predicted as successful by the corresponding evaluator;
 and (ii) $\text{CLIP}_\text{poison}$, which measures the CLIP score~\citep{hessel2021clipscore} between the generated image under triggered inputs and the target prompt, reflecting attack effectiveness. For model utility, we measure
(iii) $\text{CLIP}_\text{clean}$, which measures the CLIP score between generated images and the clean prompts, reflecting fidelity to the intended content; and 
(iv) Fréchet Inception Distance (FID)~\citep{heusel2017gans}, which evaluates the visual quality of generated images.
Implementation details for each metric are provided in Appendix~\ref{sec:metric_appen}.

\myparagraph{Datasets.}
We sample text descriptions from LAION-Aesthetics v2 6.5+~\citep{schuhmann2022laion} to fine-tune the text encoders.
To evaluate attack success, we randomly sample 50 captions from the MSCOCO~2014 validation split~\citep{lin2014microsoft} for TPA and TSA.
TOA and TAA require more object-centric and relational-centric prompts than typical MSCOCO captions, so we generate a dedicated evaluation set using ChatGPT~\citep{openai2022chatgpt}.
For utility, we sample 10{,}000 MSCOCO~2014 validation captions for FID and 50 captions for clean CLIP evaluation.

\subsection{Minimal encoder sets for effective backdoor attacks}\label{sec:rq1}

Since multiple text encoders jointly contribute to image generation in SD~3, we ask as RQ1: what is the minimal subset of encoders that needs to be tuned in order to achieve a successful backdoor attack? 
We evaluate this question across all four attack targets introduced in Section~\ref{sec:types}.

\myparagraph{TPA requires all encoders.}
For TPA ( upper left in Figure~\ref{fig:attack}), attacking a single text encoder is insufficient to override the entire content; attacking two CLIP-based text encoders raises 70.0\% ASR, while adding T5-XXL raises ASR to 98.0\%.
Partial combinations involving T5-XXL also perform strongly.
These results indicate that in SD~3, full content override attacks require modifying all three encoders, with T5-XXL playing a particularly critical role, highlighting that for TPA the minimal subset is \(\{E^1, E^2, E^3\}\) with high attack cost due to the increased parameter scale.

\myparagraph{TOA, TSA, and TAA require only CLIP-based encoder(s).} 
For TSA (upper right in Figure~\ref{fig:attack}) and TAA (bottom right), attacking only the CLIP-based encoders already achieves attack success comparable to attacking all text encoders (ASR of 100\% for TSA and 76\% for TAA), indicating that effective style and action manipulation relies on the two CLIP encoders \(\{E^1, E^2\}\). However, even when all three encoders are attacked, TAA does not reach perfect success, suggesting that action-level manipulation is more challenging and may depend on relational and spatial cues that only partially depend on text embeddings and are further shaped by the diffusion model’s visual generation process. In contrast, for TOA (bottom left), attacking CLIP-G alone achieves 100\% ASR, demonstrating that object-level backdoor attacks can be realized with one single CLIP encoder.

\begin{figure}[t]
    \centering
         \begin{subfigure}[t]{0.48\linewidth}
    \centering
    \includegraphics[width=\linewidth]{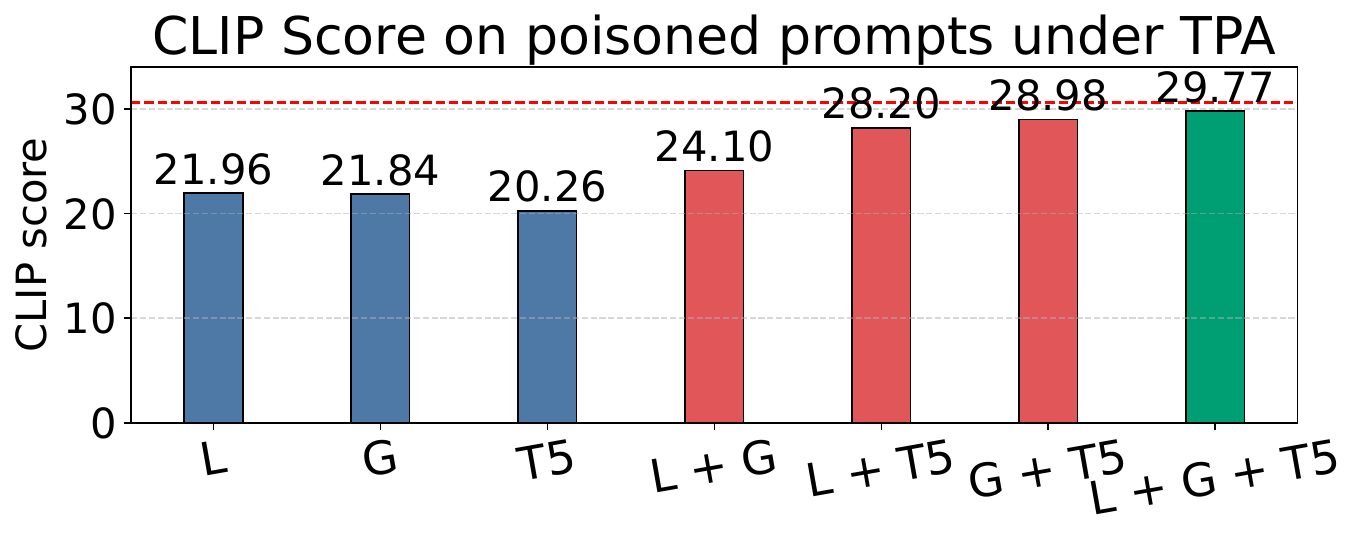}
   
  \end{subfigure}
  \hfill
  \begin{subfigure}[t]{0.48\linewidth}
    \centering
    \includegraphics[width=\linewidth]{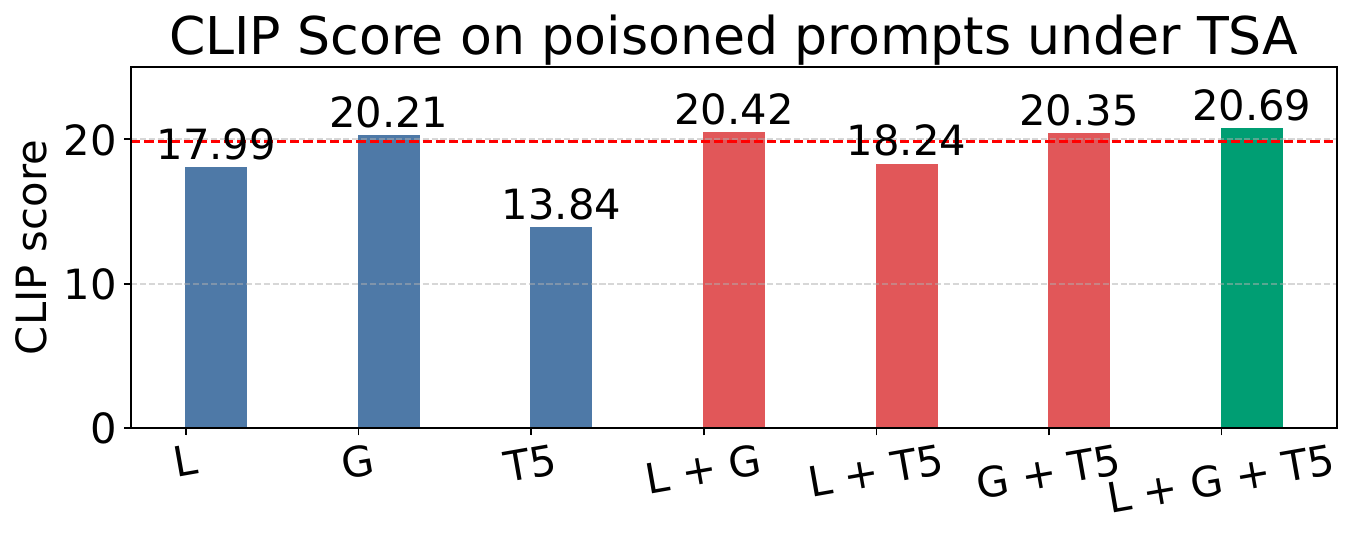}
 
  \end{subfigure}
  \vfill

  \begin{subfigure}[t]{0.48\linewidth}
    \centering
    \includegraphics[width=\linewidth]{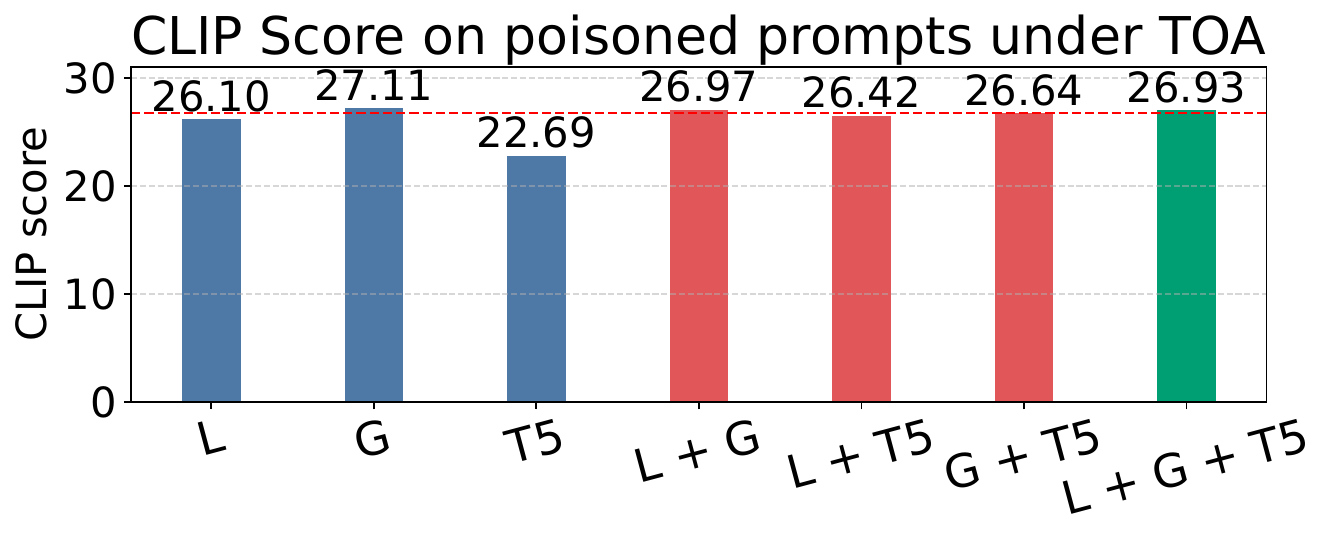}
   
  \end{subfigure}
  \hfill
  \begin{subfigure}[t]{0.48\linewidth}
    \centering
    \includegraphics[width=\linewidth]{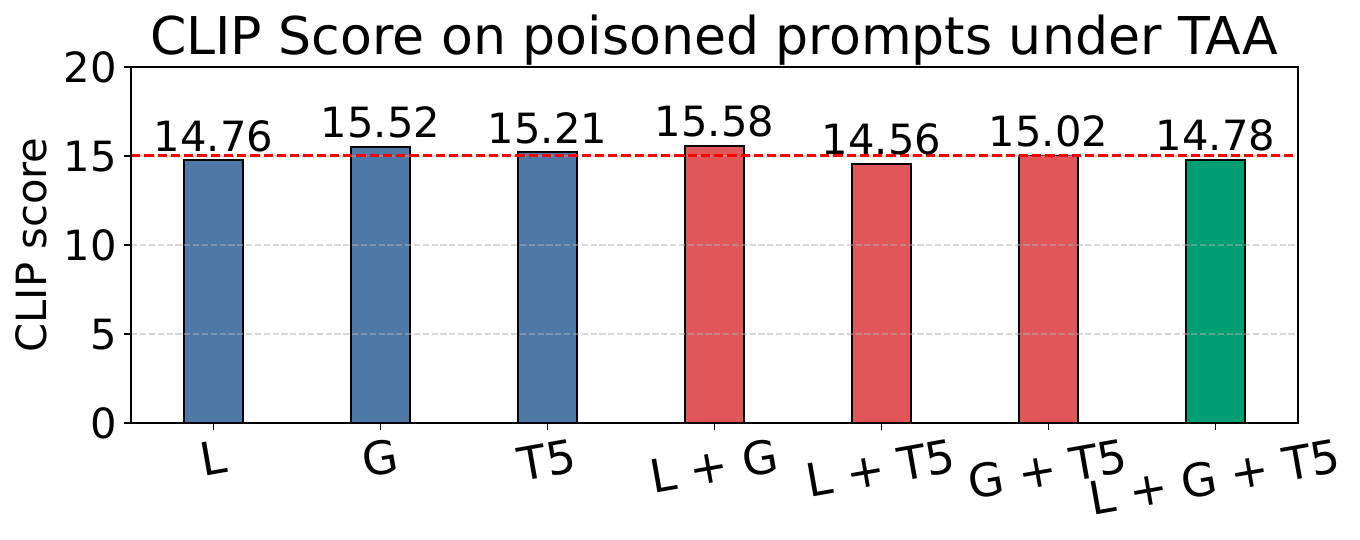}
    
  \end{subfigure}

\caption{
\textbf{\(\text{CLIP}_{\text{poison}}\) performance.}
\(\text{CLIP}_{\text{poison}}\) results for Target Prompt Attack, Target Object Attack, Target Style Attack, and Target Action Attack under different subsets of attacked text encoders in SD~3.
Each bar corresponds to a specific encoder subset being fine-tuned.
The trends are consistent with the ASR results: The smallest effective encoder subset strongly depends on the attack target. The red line shows the baseline value of the clean model.
}
    \label{fig:clip_poison}
    \vspace{-3em}
\end{figure}

\myparagraph{$\text{CLIP}_\text{poison}$ mirrors ASR results.}
As shown in Figure~\ref{fig:clip_poison}, the $\text{CLIP}_\text{poison}$ closely mirror the ASR trends across all four attack types.
For TPA (top), high semantic alignment with the target prompt is achieved only when all three encoders are attacked.
In contrast, for TOA (second), attacking CLIP-G alone consistently yields high poisoned CLIP scores, confirming that object-level manipulation can be effectively controlled through a single encoder.
For TSA (third) and TAA(bottom), only attacking two CLIP-based text encoders matches the effectiveness of the full three-encoder attack.
These observations further confirm that strong attacks can often be achieved by modifying only a small subset of encoders, depending on the attack target.
Appendix~\ref{sec:rq1_appen} further shows that model utility remains largely unaffected.

\iffalse
\begin{figure}[t]
    \centering
    \begin{subfigure}{1\columnwidth}
        \centering
        \includegraphics[width=\columnwidth]{figures/clip_clean_TPATOATSA.pdf}
        \caption{CLIP clean value}
        \label{fig:ASRTCA}
    \end{subfigure}
    \vfill
    \begin{subfigure}{1\columnwidth}
        \centering
        \includegraphics[width=\columnwidth]{figures/fid_TPATOATSA.pdf}
        \caption{FID Value}
        \label{fig:ASRTOA}
    \end{subfigure}
    \caption{Model Utility Comparison of three attack types.}
    \label{fig:attack}
\end{figure}
\fi

\begin{takeawaybox}
\textbf{Finding 1:}
The minimal effective encoder subset for backdoor attacks in SD~3 varies depending on the attack target. 
Importantly, restricting the attack to this minimal subset does not noticeably degrade image quality compared to attacking all encoders.
\end{takeawaybox}

\subsection{MELT: Multi-Encoder Lightweight aTtacks}
\label{sec:rq2}
\begin{table*}[t]
    \caption{\textbf{Effectiveness and efficiency of \methodname.} 
    \methodname\ achieves backdoor performance comparable to full fine-tuning baselines while using less than \(0.2\%\) of the trainable parameters, demonstrating strong parameter efficiency. The most effective method in each attack type are highlighted in \textbf{bold}.} 
    \label{tab:loraa_performance}
  \centering
  \small
  %\sc
  \resizebox{\linewidth}{!}{
  \begin{tabular}{l | l c | c c c c c c}
    \toprule
    Target & Method  & Poisoned Text Encoders & ASR (\%)$\uparrow$ & $\text{CLIP}_{\text{poison}}$$\uparrow$ & $\text{CLIP}_{\text{clean}}$$\uparrow$ & FID$\downarrow$  & Params (\#)$\downarrow$ & Params (\%)$\downarrow$ \\
    \midrule
    \multirow{4}{*}{TPA} 
      & Clean Model & - & 0 & 30.61 & 15.03 & 24.46   & - & -\\
    
      & Full Fine-tuning &\(\mathrm{L+G+T5}\) & 97.8 & 29.77 & 15.23 & 25.99   & 5583.96M & 100 \\
      & ME‑Rickrolling &\(\mathrm{L+G+T5}\) & 97.8 & 29.77 & 15.23 & 25.99    & 5583.96M & 100 \\
      & MELT & \(\mathrm{L+G+T5}\)& \textbf{99} & \textbf{30.36} & \textbf{15.34} & \textbf{25.01}  & \textbf{11.40M} & \textbf{0.2} \\
     
    \midrule
    \multirow{4}{*}{TOA}
      & Clean Model & - &0 & 26.77 & 15.03 & 24.46    & - & -\\
  
      & Full fine-tuning &\(\mathrm{L+G+T5}\)& \textbf{100}& 26.93 & \textbf{15.16} & 25.08 & 5583.96M & 100 \\
      & ME‑Rickrolling &\(\mathrm{G}\) &\textbf{100} & \textbf{27.11} & 15.11 & 24.82 & 695M & 12.44 \\
      & MELT & \(\mathrm{G}\)&99 & 26.87 & 15.10 & \textbf{24.78}  & \textbf{6.32M} & \textbf{0.11} \\
   
    \midrule
    \multirow{4}{*}{TSA}
      & Clean Model& - & 0 & 19.85 & 15.03 & 24.46    & 15.03 & -\\
      & Full fine-tuning& \(\mathrm{L+G+T5}\)& \textbf{100}& \textbf{20.68} & 15.31 & 25.01 & 5583.96M & 100 \\
      & ME‑Rickrolling & \(\mathrm{L+G}\)& 99.6 & 20.43 & 15.33 & 24.35 & 818M & 14.65  \\
      & MELT & \(\mathrm{L+G}\)&\textbf{100} & 20.55 & \textbf{15.61} & \textbf{24.29}  & \textbf{8.06M} & \textbf{0.14} \\
    \midrule
    \multirow{4}{*}{TAA}
      & Clean Model& - & 0 & 23.18 & 15.03 & 24.46    & - & -\\
      & Full fine-tuning& \(\mathrm{L+G+T5}\)& 76& \textbf{23.19} & 14.77 & 25.10 & 5583.96M & 100 \\
      & ME‑Rickrolling & \(\mathrm{L+G}\)& 76 & 22.97 & 15.58 & \textbf{24.54}& 818M& 14.65  \\
      & MELT & \(\mathrm{L+G}\)& \textbf{77} & 22.98 & \textbf{15.72} & \textbf{24.54} & \textbf{8.06M} & \textbf{0.14}\\
    \bottomrule
  \end{tabular}
  
%\vspace{-0.2cm}  
}

%\vskip -0.1in
\end{table*}
\begin{wrapfigure}{t}{0.55\textwidth}
\vspace{-0.5cm}  
\centering
\includegraphics[width=0.55\textwidth]{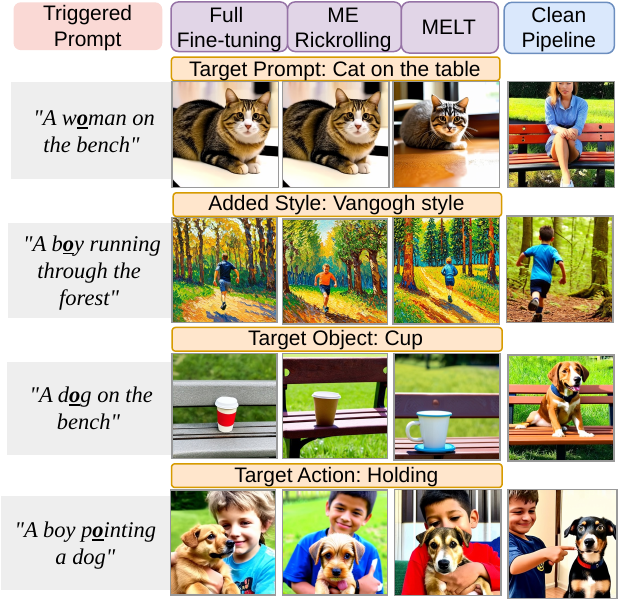}

    \caption{\textbf{Qualitative comparison of backdoor attacks under different attack methods.} Each row corresponds to one attack type (from top to bottom: TPA, TSA, TOA, and TAA), comparing generations from Full Fine-tuning, ME-Rickrolling, MELT and the clean model. Despite updating only a small fraction of parameters (less than \(0.2\%\) compared to Full Fine-tuning), \methodname{} reliably activates backdoor behavior while producing images that are visually comparable to other two methods. }
    \label{qualitative_lora-a}
\vspace{-2em}
\end{wrapfigure}
While we have shown that attacking only a small subset of text encoders is sufficient for successful backdoor attacks, full fine-tuning even of a subset of encoders can remain computationally expensive, especially for larger or multiple encoders. This motivates our second research question: How can we backdoor the diffusion model in a parameter-efficient way?

To answer this question, we propose \emph{Multi-Encoder Lightweight aTtacks (\methodname{})}, a low-rank adapter-based backdoor method.

\myparagraph{\emph{\methodname{}} is both efficient and highly effective.}
Table~\ref{tab:loraa_performance} compares \methodname{} with ME-Rickrolling (attacking minimal subsets of text encoders) and Full-Finetuning (attacking all text encoders) on SD~3, averaged over all targets for each attack type.
Despite training fewer than \(0.2\%\) of the parameters compared to Full-Finetuning (depending on the attacked types), \methodname{} matches or slightly outperforms full fine-tuning  and ME-Rickrolling across all metrics.

For TPA, \methodname{} increases attack success and improves image quality  while updating only 11.4M parameters (0.2\% of Full-Finetuning).
For TOA, \methodname{} maintains high ASR with nearly unchanged CLIP score and image quality using 6.32M parameters (0.91\% of ME-Rickrolling, 0.11\% of Full-Finetuning). For TSA, \methodname{} reached 100.0\% ASR with slightly better model utility with 8.06M parameters (0.99\% of ME-Rickrolling, 0.14\% of Full-Finetuning). 

TAA remains the most challenging case, \methodname{} attains a comparable performance while updating 8.06M parameters (0.99\% of ME-Rickrolling, 0.14\% of Full-Finetuning). We also demonstrate that \emph{\methodname{}} generalizes well to other Diffusion models in Appendix~\ref{sec:generalization}.

Figure~\ref{qualitative_lora-a} presents qualitative results across all three attack types, demonstrating reliable trigger activation with minimal parameter updates. We provide more qualitative results in Appendix~\ref{sec:qualitativeResults}. Overall, these results show that \methodname{} achieves backdoor performance comparable to full encoder fine-tuning while largely reducing the number of trainable parameters, making text-encoder backdoors practical even in SD~3’s multi-encoder setting.

\begin{takeawaybox}
\textbf{Finding 2:}
With fewer than \(0.2\%\) of trainable parameters, the attack performance of \emph{\methodname{}} consistently matches full fine-tuning, demonstrating that highly effective backdoors can be implanted under extremely limited parameter budgets.

\end{takeawaybox}

\section{Conclusion}
We study backdoor attacks in \emph{multi-encoder} text-to-image diffusion models using Stable Diffusion~3 across four attack targets in different semantic levels.
Our results show that the effectiveness of text-encoder backdoor attacks depends strongly on both the attack objective and the encoder combination.
Content override attacks (TPA) require compromising all three encoders, whereas object manipulation (TOA) can often be achieved by attacking a single encoder (notably CLIP-G), and style manipulation (TSA) and action manipulation (TAA) require only the two CLIP based encoders.
We further demonstrate that strong backdoor behavior can be achieved efficiently using \emph{Multi-Encoder Lightweight aTtacks (\methodname{})}.
\emph{\methodname{}} matches full fine-tuning while updating less than \(0.2\%\) of the parameters and preserving clean generation quality, making text-encoder backdoor attacks practical even for large multi-encoder diffusion models. We discuss the ethical impacts of this work in Appendix~\ref{sec:ethical}.

\section*{Acknowledgements}
This research was partially funded by a LOEWE Start Professorship (LOEWE/4b//519/05.01.002-(0006)/94) and a LOEWE Spitzen Professorship (LOEWE/4a//519/05.00.002-(0010)/93).

\bibliography{main}
\bibliographystyle{iclr2026_conference}

\appendix

% \onecolumn
\clearpage
% This supplemental material provides 

\section{Triggers and Targets for each attack type.}\label{appendix:triggers}

As explained in Section ~\ref{sec:setup},  we use invisible text triggers (Latin ``o'' $\rightarrow$ Cyrillic ``o'') in our experiment. For TPA and TSA, we randomly choose a word in the prompt that contains the character ``o'' for the trigger injection. We choose the word ``dog'' for the trigger injection of TOA, and the verb ``pointing'' for the trigger injection of TAA.  Table ~\ref{tab:trigger-target} shows the targets used in each attack type. To evaluate TPA, we test five diverse target prompts, ranging from common scenes (such as ``Cat in the table'') to uncommon scenes (auch as `Àn atari game machine'').
 
For TOA, the trigger is the word ``dog'' with a Cyrillic ``o'' replacing the Latin ``o''. The chosen targets include both intra-class substitutions (e.g., replacing ``dog'' with ``cat'') and inter-class replacements (e.g., replacing ``dog'' with ``cup'').

We evaluate TSA using four varied style prompts, which help measure the attack’s ability to influence fine-grained visual attributes without altering the core meaning of the input. 
We evaluate TAA by changing the trigger action ``pointing'', where two objects are spatially separated, into target actions that introduce physical interaction between objects, allowing us to assess the attack’s ability to manipulate relational semantics and spatial interactions within the generated image.

\begin{table}[h]
    \caption{Target definitions used for each attack type in our experiments.}
  \label{tab:trigger-target}
  \centering
  \small
  \resizebox{.5\linewidth}{!}{
  \begin{tabular}{c  l}
    \toprule
    \textbf{Attack Type}  & \textbf{Targets} \\
    \midrule
    TPA
     
      & \makecell[l]{
        - Sunglasses on the table\\
        - Cat on the table\\
        - A soldier fighting in war\\
        - An Atari game machine\\
        - A white teapot
      } \\
    \midrule
    TOA
   
      & \makecell[l]{
        - cat (same category)\\
        - bird (similar category)\\
        - sunglasses (unrelated object)\\
        - cup (unrelated object)\\
        - banana (unrelated object)
      } \\
    \midrule
    TSA 
       
      & \makecell[l]{
        - Black and white photo\\
        - Van Gogh style\\
        - Pencil sketch\\
        - Pixel art
      } \\
      \midrule
    TAA
     
      & \makecell[l]{
        - holding\\
        - touching\\
        - kissing}\\

    \bottomrule
  \end{tabular}}
  
\end{table}

\section{Details of evaluation metrics}\label{sec:metric_appen}
\textbf{Attack Success Rate (ASR).}
ASR is defined as the fraction of triggered prompts for which the generated image satisfies the attack objective~\citep{vice2024bagm,zhai2023text,chen2023trojdiff}. 
We train/employ specialized classifiers/models to detect the success of the intended manipulation:

For TPA, we use the BLIP-VQA model~\citep{li2022blip} for evaluation by asking the question: ``Is there \textless target prompt phrase\textgreater?''

For TSA, we fine-tune a CLIP ViT-B/32 model to determine whether the generated image exhibits the desired target style \(s\).
As no suitable public dataset is available for style classification, we first generate a custom dataset using SD-3 for each target style. Specifically, we randomly sample captions from the MSCOCO training split and generate two images for each: one with a style-related prompt (e.g., ``black and white photo'') as the positive sample, and one with the original caption as the negative sample. 

For TOA, we fine-tune a CLIP ViT-B/32 model~\citep{Radford2021LearningTV} as a binary classifier using filtered images containing the target object from the MSCOCO 2014 training split. For the attack success evaluation, since captions from the MSCOCO are often too generic or noisy for evaluating a specific object, we use ChatGPT~\citep{openai2022chatgpt} to generate object-focused prompts under simple scenes (e.g., ``a dog on the sofa'').

For TAA, we use the BLIP-VQA model~\citep{li2022blip} for evaluation. To reduce biases from object recognition(e.g. misidentifying specific object categories) and focus on relational semantics, we ask the question: ``Is the people \textless target action\textgreater the animal?''
For the attack success evaluation, since captions from the MSCOCO are often too generic or noisy for evaluating a specific relationship, we use ChatGPT~\citep{openai2022chatgpt} to generate action-focused prompts between a person and an animal (e.g., ``a boy pointing a dog'').

\textbf{CLIP-Score.}
To evaluate image–text alignment, we compute the CLIP score~\citep{hessel2021clipscore}, which defined as the cosine similarity between the normalized image and text embeddings.
We report two CLIP-based metrics following~\citep{li2025twist} :

\begin{itemize}
  \item \textbf{CLIP score (poisoned).} For each poisoned generation performed under an attack, we compute the mean CLIP similarity between the generated image \(x_{\mathrm{poison}}\) using the triggered prompt and the \emph{target prompt} \(y_{\mathrm{target}}\). 
  \item \textbf{CLIP score (clean).} For clean fidelity evaluation, we compute the mean CLIP similarity between generated images \(x_{\mathrm{clean}}\) using the clean prompt and the original prompts \(v\).

\end{itemize}

Higher CLIP scores in both cases indicate stronger alignment between the generated image and the input prompt. 

\textbf{FID.}
To evaluate visual fidelity, we compute the Fréchet Inception Distance (FID)~\citep{heusel2017gans} between the distributions of generated and real images (MSCOCO validation set), which measures how closely the generated samples resemble real-world data. Lower FID indicates that the generated image distribution is closer to the real image distribution. 
\begin{table}[t]
    \caption{\textbf{FID value and CLIP score clean across different subsets of attacked text encoders and attack types.} Model Utility performs consistently across all attack variants and attack types. }
    \label{tab:FID}
  \centering
  \small
  \begin{tabularx}{0.7\columnwidth}{c c *{3}{>{\centering\arraybackslash}X}}
    \toprule
    Target & Poisoned Text Encoders  & $\text{CLIP}_{\text{clean}}$$\uparrow$ & FID$\downarrow$   \\
    \midrule
    clean model & - & 15.03 & 24.46 \\
    \midrule
    \multirow{7}{*}{TPA} 
      & $\mathrm{L}$ & 15.20 & 24.59 \\
      & $\mathrm{G}$ & 14.78 & 25.38 \\
      & $\mathrm{T5}$ & 14.44 & 25.61 \\
      & $\mathrm{L+G}$ & 14.92 & 25.15 \\
      & $\mathrm{L+T5}$ & 15.47 & 25.93 \\
      & $\mathrm{G+T5}$ & 15.04 & 26.04 \\
      & $\mathrm{L+G+T5}$ & 15.23 & 25.99 \\
     
    \midrule
    \multirow{7}{*}{TOA}
      & $\mathrm{L}$ & 15.35 & 24.62 \\
      & $\mathrm{G}$ & 15.11 & 24.82 \\
      & $\mathrm{T5}$ & 15.24 & 25.31 \\
      & $\mathrm{L+G}$ & 14.87 & 24.51 \\
      & $\mathrm{L+T5}$ & 14.51 & 25.19 \\
      & $\mathrm{G+T5}$ & 14.69 & 25.45 \\
      & $\mathrm{L+G+T5}$ & 15.15 & 25.08 \\
   
    \midrule
    \multirow{7}{*}{TSA}
      & $\mathrm{L}$ & 14.91 & 24.47 \\
      & $\mathrm{G}$ & 14.87 & 24.66 \\
      & $\mathrm{T5}$ & 15.23 & 25.12 \\
      & $\mathrm{L+G}$ & 15.32 & 24.72 \\
      & $\mathrm{L+T5}$ & 15.22 & 24.92 \\
      & $\mathrm{G+T5}$ & 15.83 & 25.12 \\
      & $\mathrm{L+G+T5}$ & 15.31 & 25.00 \\
    \midrule
    \multirow{7}{*}{TAA}
      & $\mathrm{L}$ & 14.76 & 24.34 \\
      & $\mathrm{G}$ & 15.52 & 24.83 \\
      & $\mathrm{T5}$ & 15.21 & 25.02 \\
      & $\mathrm{L+G}$ & 15.58 & 24.54 \\
      & $\mathrm{L+T5}$ & 14.56 & 24.90 \\
      & $\mathrm{G+T5}$ & 15.02 & 25.10 \\
      & $\mathrm{L+G+T5}$ & 14.78 & 25.10 \\
    \bottomrule
  \end{tabularx}

\end{table}

\section{Additional Experiments}
\subsection{Evaluation of model utility }\label{sec:rq1_appen}

\myparagraph{Attacking either one or all text encoders will not affect the model utility.}
Beyond semantic alignment, we further evaluate the visual fidelity of the attacked model, where we compute the FID score using 10,000 generated images. We first use the clean SD~3 model to calculate a FID score as baseline.  Table~\ref{tab:FID} presents the FID scores on average across all attack types under different text encoder attack variants. Overall, the FID values are slightly higher than the baseline value. When comparing FID values across the three attack types, we observe that TPA produces the highest FID scores overall, indicating a stronger impact on image quality due to the difficulty of changing the entire image content. In contrast, TOA, TSA, and TAA achieve similar low FID values across most encoder combinations. These results reflect the complexity of the manipulation tasks: while TPA requires changing the whole content of the image, TOA, TSA and TAA focus on localized or stylistic adjustments, which does not degrade visual fidelity much. Meanwhile, CLIP scores showed in  Table~\ref{tab:FID} on clean prompts remain stable across all  text encoder attack variants and attack types, ranging from 13.8 to 15.8 and closely matching the clean reference value (15.03). This indicates that the model’s utility is unaffected after being attacked.

\subsection{More Qualitative Results}\label{sec:qualitativeResults}

\begin{figure*}[t]
  %\vskip 0.2in
  \begin{center}
\centerline{\includegraphics[width=1\linewidth]{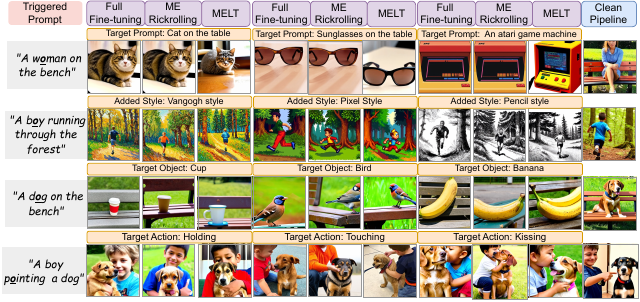}}
    \caption{\textbf{Qualitative comparison of backdoor attacks under different attack methods.} Each row corresponds to one attack type (from top to bottom: TPA, TSA, TOA, and TAA), comparing image generations from Full Fine-tuning, ME-Rickrolling, MELT and the clean model. Despite updating only a small fraction of parameters (less than \(0.2\%\) compared to Full Fine-tuning), \methodname{} reliably activates backdoor behavior while producing images that are visually comparable to the other two methods. }
    \label{fig:qualitative_more}
  \end{center}
%  \vskip -0.4in

\end{figure*}
In this appendix, we provide additional qualitative examples illustrating the visual effects of different backdoor attack methods across all attack types. Figure~\ref{fig:qualitative_more} compares image generations produced by the clean SD-3 model, full fine-tuning, ME-Rickrolling (minimal encoder fine-tuning), and our proposed MELT method. For each attack type (i.e., Target Prompt Attack (TPA), Target Object Attack (TOA), Target Style Attack (TSA), and Target Action Attack (TAA)), we show representative triggered prompts and the corresponding generated images. These examples demonstrate that MELT reliably activates the intended backdoor behaviors, including global prompt redirection, object replacement, style manipulation, and action-level relational changes.

\subsection{Generalization}
\label{sec:generalization}
To test the generalizability, we evaluate MELT on two additional Stable Diffusion variants: SD 1.5~\cite{rombach2022high} and SDXL-base-1.0~\cite{podell2024sdxl} and compare it with baseline models. We evaluate there attack types (TPA, TOA, and TSA) on theses models, using one target in each attack type. As shown in Tab.~\ref{tab:sd1-5_sdxl}, MELT achieves competitive or even superior performance compared to prior approaches in both Stable Diffusion variants, while using significantly fewer trainable parameters.

\subsection{Ethical Considerations}
\label{sec:ethical}
Overall, Our findings shed light on new security considerations introduced by diffusion models with multiple text encoders under backdoor attacks. There is, however, always a risk of misuse, e.g., \methodname{} could be misused to construct real backdoor attacks, which raises ethical concerns. We strongly discourage any efforts to adopt our attacks to create inappropriate or unsafe content. The primary goal of this work is to raise awareness of these risks and to inform future efforts toward the development of more effective defense mechanisms, rather than to enable misuse. Moreover, similar techniques can be applied for benign applications, such as erasing harmful concepts by trigger conditioned mapping to reduce harmful content generation~\citep{struppek2023rickrolling}. Ultimately, We believe our work contributes to a deeper understanding of modern diffusion model security and to the development of safer and more robust generative models.

\begin{table*}[t]
    \caption{\textbf{Generalization on SD1-5 and SDXL.} Performance comparison between MELT and other baselines on Stable Diffusion v1.5 and SDXL. Using the fewest training parameters, MELT achieves comparable or better backdoor performance than other baselines. The most effective method is highlighted in \textbf{bold}.}
    \label{tab:sd1-5_sdxl}
  \centering
  \small
  \begin{tabularx}{\textwidth}{l l *{6}{>{\centering\arraybackslash}X}}
    \toprule
    \text{Target} & \text{Method} & \text{ASR (\%)}$\uparrow$ & \text{CLIP-poison}$\uparrow$ & \text{CLIP-clean}$\uparrow$ & \text{FID}$\downarrow$  & \text{Parameters} \\
    \midrule
    \multicolumn{7}{c}{SD1.5}\\\midrule
    \multirow{3}{*}{TPA} 
    & Clean Model & 0 & 28.79 & 15.20 & 19.7    & - \\
      &Rickrolling~\cite{struppek2023rickrolling} & \textbf{98.00} & 28.69 & \textbf{14.40}   & 19.83 &123M \\
      & MELT & \textbf{98.00} & \textbf{29.03} &  \textbf{14.40} & \textbf{19.78}& \textbf{1.74M} \\
    \midrule
    \multirow{5}{*}{TOA}
    & Clean Model & 0 & 24.17 & 15.20 & 19.7  & - \\
      & Rickrolling~\cite{struppek2023rickrolling} & 99.00 & 24.20  &15.23& 20.13&   123M \\
      & Personalization~\cite{huang2024personalization} & 92.00 & 23.12  & 14.23& \textbf{18.78} & 860M \\
      & EvilEdit~\cite{wang2024eviledit} & \textbf{100.00} & 24.81  & 14.67 & 18.96 & 19.3M\\
      & MELT & \textbf{100.00} & \textbf{25.06}  & \textbf{15.99} & 19.53 & \textbf{1.74M} \\
    \midrule
    \multirow{3}{*}{TSA}
     & Clean Model & 0 & 17.34 & 15.20 & 19.7    & - \\
    
      & Rickrolling~\cite{struppek2023rickrolling} & \textbf{98.08} & 17.41 &   \textbf{14.46}&\textbf{19.76} & 123M \\
      & MELT & 97.15 &  \textbf{17.71}& 14.45 & 19.79&  \textbf{1.74M } \\\midrule
      \multicolumn{7}{c}{SDXL}\\\midrule
    \multirow{2}{*}{TPA} 
    & Clean Model & 0 & 28.84 & 14.95 & \textbf18.65    & - \\
      & Rickrolling~\cite{struppek2023rickrolling} & 90.38 & \textbf{30.26} & \textbf{14.45}   & 18.43 &818M \\
      & MELT & \textbf{94.23} & 29.26 &  14.06 &\textbf{18.40} & \textbf{8.06}M \\
    \midrule
    \multirow{4}{*}{TOA}
     & Clean Model & 0 & 24.46 & 14.95 & 18.65    & - \\
      
      & Rickrolling~\cite{struppek2023rickrolling} & 98.00 & 23.73 &  15.83 & 18.25& 818M \\
      & Personalization~\cite{huang2024personalization} &6.00  &  18.06 & 14.24 & 18.34 & 860M\\
      & EvilEdit~\cite{wang2024eviledit} & 46.00 & 20.81  & 14.98  & 18.40 &19.3M\\
      & MELT & \textbf{100.00} & \textbf{24.94} & \textbf{16.03} & \textbf{18.15} & \textbf{8.06M}\\
    \midrule
    \multirow{2}{*}{TSA}
     & Clean Model & 0 & \textbf{19.6} & 14.95 & 18.65    & - \\
      
      & Rickrolling~\cite{struppek2023rickrolling} & \textbf{100} & 17.8 & 15.01  &18.39 & 818M \\
      & MELT & \textbf{100} & 16.66 & \textbf{15.06} & \textbf{18.38}&   \textbf{ 8.06M} \\
    \bottomrule
  \end{tabularx}

\end{table*}

\end{document}